\documentclass[11pt]{article}

\usepackage[margin=1in]{geometry}
\usepackage[T1]{fontenc}
\usepackage{lmodern}
\usepackage{microtype}
\usepackage{amsmath,amssymb}
\usepackage{booktabs}
\usepackage{graphicx}
\usepackage{tikz}
\usetikzlibrary{arrows.meta,fit}
\usepackage[numbers]{natbib}
\usepackage{hyperref}
\usepackage{xcolor}

\hypersetup{
  colorlinks=true,
  linkcolor=blue,
  citecolor=blue,
  urlcolor=blue
}

\title{A Multi-Dimensional Quality Scoring Framework for Decentralized LLM Inference with Proof of Quality}
\author{
  Arther Tian\textsuperscript{a},
  Alex Ding\textsuperscript{a,*},
  Frank Chen\textsuperscript{a}\\
  Simon Wu\textsuperscript{a},
  Aaron Chan\textsuperscript{a}\\
  \textsuperscript{a}DGrid AI\\[0.5em]
  \textsuperscript{*}Corresponding author: \texttt{alex.ding@dgrid.ai}
}

\date{}

\begin{document}
\maketitle

\begin{abstract}
Decentralized large language model (LLM) inference networks can pool heterogeneous compute to scale serving, but they require lightweight and incentive-compatible mechanisms to assess output quality.
Prior work introduced cost-aware Proof of Quality (PoQ) and adaptive robust PoQ to allocate rewards under evaluator heterogeneity and adversarial behavior.
In this paper, we focus on the quality signal itself and propose a multi-dimensional quality scoring framework that decomposes output quality into modular dimensions, including model and cost priors, structure quality, semantic quality, query-output alignment, and agreement/uncertainty.
Using logged outputs from QA and summarization tasks, we systematically audit dimension reliability and show that seemingly reasonable dimensions can be task-dependent and even negatively correlated with reference quality without calibration.
While the default composite underperforms a strong single semantic evaluator, ablations reveal that removing unreliable dimensions and re-normalizing weights yields a calibrated composite that matches or exceeds the best single-evaluator and consensus baselines.
Finally, we integrate the composite score as a drop-in quality signal in PoQ and demonstrate complementary benefits with robust aggregation and adaptive trust weighting under adversarial evaluator attacks.
\end{abstract}

\section{Introduction}
\label{sec:intro}

Decentralized LLM inference has emerged as a practical direction to meet growing demand under constrained and heterogeneous compute.
Systems that enable collaborative inference across distributed participants illustrate the feasibility of pooling resources at scale, while modern serving optimizations highlight that end-to-end throughput and memory efficiency remain central bottlenecks even in centralized settings \citep{borzunov-etal-2023-petals,kwon2023pagedattention,dao2022flashattention}.
A fundamental challenge in decentralized inference is \emph{verifying and pricing quality}: participants may contribute different models, hardware, and serving policies, and the network must assign rewards that reflect the usefulness of produced outputs.

Proof of Quality (PoQ) offers a lightweight alternative to heavy cryptographic verification by relying on evaluator models (or learned metrics) to score outputs and drive consensus and incentives.
Our prior work developed cost-aware PoQ to jointly consider output quality and evaluation cost, and further proposed adaptive robust PoQ mechanisms to withstand malicious evaluators and strategic manipulation \citep{tian2025costawarepoq,tian2026adaptiverobustpoq}.
However, these approaches inherit a critical dependency: \emph{the quality signal itself} must be reliable across tasks and robust to evaluator bias.
In practice, automatic evaluation metrics exhibit significant variance, and even strong learned metrics can fail under distribution shifts or task-specific criteria \citep{lin-2004-rouge,zhang2020bertscore,sellam-etal-2020-bleurt,liang2023helm}.

This paper argues that decentralized inference should move beyond single-evaluator scoring toward a \emph{multi-dimensional} view of quality.
Different signals capture different failure modes: structural heuristics can detect degeneration and formatting violations; semantic metrics capture meaning preservation; instruction/query-output alignment targets compliance; and agreement or uncertainty across evaluators can reflect confidence.
Yet, combining multiple signals is not automatically beneficial: some dimensions may be task-dependent, correlated with spurious features, or even negatively aligned with human preferences if used na\"ively.
Therefore, the central problem becomes \emph{dimension design and reliability}: which dimensions are consistently informative, when do they fail, and how should they be calibrated before being used for incentives?

\begin{figure}[t]
  \centering
  \includegraphics[width=0.98\linewidth]{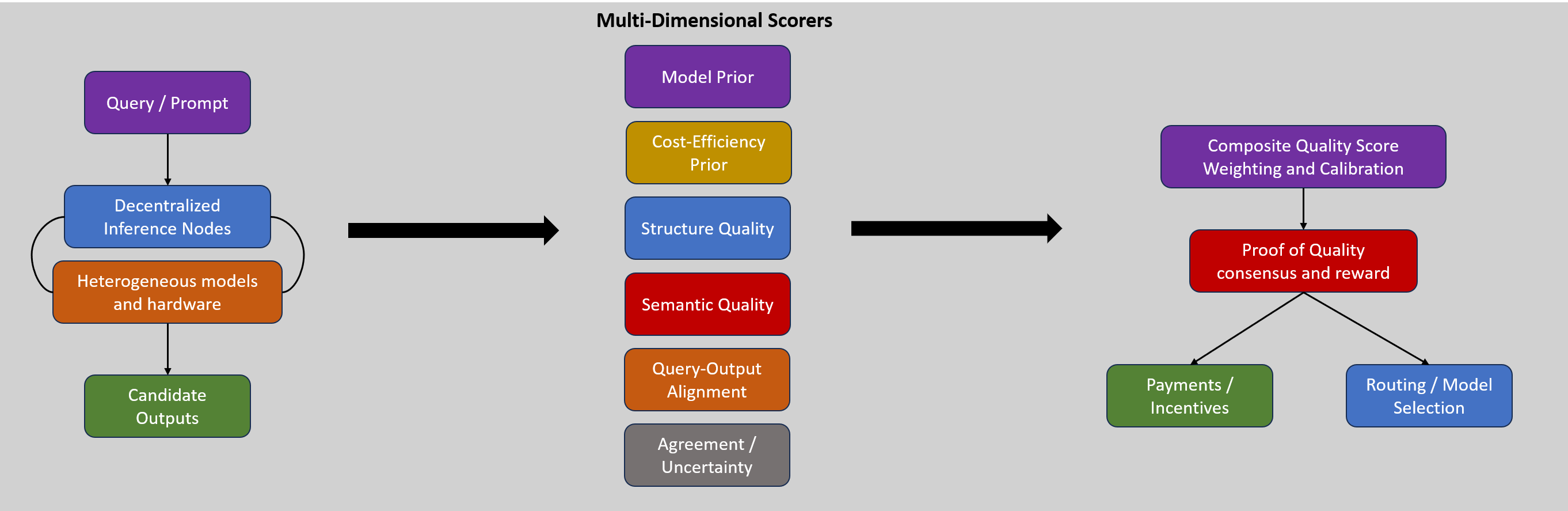}
  \caption{Overview of the proposed multi-dimensional quality scoring framework and its integration into Proof of Quality (PoQ) for decentralized LLM inference. Candidate outputs are scored by multiple dimension modules and combined into a composite quality signal that can be used for consensus and rewards.}
  \label{fig:mdqs-overview}
\end{figure}

Figure~\ref{fig:mdqs-overview} illustrates our framing: a multi-dimensional quality scorer sits between decentralized generation and PoQ consensus, producing a composite signal that can be audited, calibrated, and swapped as tasks evolve.
This design complements the broader trend of holistic evaluation and human-preference-based benchmarking \citep{liang2023helm,zheng2023judging,chiang2024chatbotarena}.

\paragraph{Contributions.}
\begin{itemize}
  \item We propose a \textbf{multi-dimensional quality scoring framework} for decentralized LLM inference, organizing quality signals into interpretable dimension modules and composing them into a composite score.
  \item We provide a \textbf{systematic reliability analysis} of dimensions and composites across tasks, including correlation with reference signals, ablations, and failure-mode characterization.
  \item We demonstrate \textbf{PoQ integration} by treating the composite score as a drop-in quality module for decentralized consensus and reward allocation, connecting quality-signal design to incentive mechanisms \citep{tian2025costawarepoq,tian2026adaptiverobustpoq}.
\end{itemize}

\paragraph{Paper organization.}
Section~\ref{sec:background} reviews PoQ and the challenges of quality-signal design.
Section~\ref{sec:framework} presents the multi-dimensional scoring framework.
Sections~\ref{sec:method} and \ref{sec:results} describe experimental methodology and results.
Section~\ref{sec:poq-integration} discusses PoQ integration, followed by related work and conclusions.

\section{Background and Problem Setting}
\label{sec:background}

This section provides the minimal background needed to situate multi-dimensional quality scoring in decentralized LLM inference networks.
We first summarize Proof of Quality (PoQ) as a lightweight verification-and-incentive mechanism, then articulate why \emph{quality signal design} becomes the central bottleneck once cryptographic verification is impractical at scale.

\subsection{Proof of Quality for Decentralized LLM Inference}
\label{subsec:poq}

\paragraph{System setting.}
We consider a decentralized inference network in which a set of inference nodes execute LLM inference and return candidate outputs for user queries.
A separate set of evaluator nodes (or evaluation services) produce scores for these outputs.
The network aggregates evaluation outcomes to drive (i) consensus on which outputs are considered high-quality, and (ii) reward allocation to inference nodes.
This setup reflects a practical trade-off: directly verifying inference correctness via cryptographic proofs remains costly and complex for large models and real-time serving, despite major advances in verifiable computation and succinct proofs \citep{parno2013pinocchio,bensasson2014vonneumann}.

\paragraph{PoQ intuition.}
PoQ replaces heavyweight verification with \emph{statistical} or \emph{learned} evaluation.
Instead of proving execution correctness, PoQ aims to provide an incentive-compatible approximation to ``quality'' using evaluator scores, while explicitly accounting for evaluation cost and adversarial risk \citep{tian2025costawarepoq,tian2026adaptiverobustpoq}.
In decentralized environments, evaluator heterogeneity (different evaluator models/metrics, different cost/latency profiles) is the norm, which motivates cost-aware sampling and robust aggregation.

\paragraph{Consensus and robustness.}
Since evaluators can be noisy or malicious, PoQ naturally interacts with robust statistics and Byzantine-resilient aggregation principles.
Classic BFT-style consensus highlights the need for tolerating adversarial components in distributed decision-making \citep{castro1999pbft}.
Similarly, Byzantine-robust learning shows that naive averaging can be fragile under targeted attacks, and motivates trimmed/median-style or adaptive trust-weight approaches \citep{blanchard2017krum,pmlr-v80-yin18a,mhamdi2018hidden}.
Our prior PoQ work adopted these principles to reduce sensitivity to malicious evaluators while preserving incentive alignment \citep{tian2026adaptiverobustpoq}.

\begin{table}[t]
  \centering
  \caption{Notation and key components in decentralized inference with PoQ.}
  \label{tab:notation}
  \begin{tabular}{@{}ll@{}}
    \toprule
    Symbol / Term & Meaning \\
    \midrule
    $q$ & User query / prompt \\
    $i \in \mathcal{I}$ & Inference node index \\
    $y_i$ & Candidate output produced by inference node $i$ \\
    $e \in \mathcal{E}$ & Evaluator / metric index \\
    $s_e(y_i, q)$ & Score from evaluator $e$ for output $y_i$ given query $q$ \\
    $\hat{s}(y_i, q)$ & Aggregated score used by consensus / rewards \\
    $c_e$ & Cost (latency, \$, compute) of running evaluator $e$ \\
    $\pi(i)$ & Reward/payment assigned to inference node $i$ \\
    \bottomrule
  \end{tabular}
\end{table}

\subsection{Quality Signal Design Challenges}
\label{subsec:quality-challenges}

\paragraph{Why quality signals dominate system behavior.}
Once PoQ (or any evaluator-driven mechanism) is adopted, the \emph{choice and design of the quality signal} becomes the main determinant of reward ranking and, therefore, participant incentives.
If the quality signal is misaligned with human preference or task objectives, rational participants may optimize toward the evaluator rather than the user.
This phenomenon is well known in text generation evaluation: surface-overlap metrics such as BLEU/ROUGE can be brittle or gameable, while learned metrics can embed biases and fail under distribution shift \citep{papineni-etal-2002-bleu,lin-2004-rouge,zhang2020bertscore,sellam-etal-2020-bleurt,rei-etal-2020-comet}.

\paragraph{Task dependence and metric mismatch.}
Different tasks emphasize different notions of quality.
Summarization requires faithfulness and factual consistency beyond semantic similarity, motivating dedicated evaluators and consistency checks \citep{kryscinski-etal-2020-evaluating,scialom-etal-2021-questeval,laban-etal-2022-summac}.
For instruction-following and open-ended dialogue, holistic and preference-based evaluation protocols have emerged, but remain imperfect and expensive to scale \citep{liang2023helm,zheng2023judging,chiang2024chatbotarena}.
Consequently, a single evaluator is rarely sufficient across diverse task distributions.

\paragraph{Evaluator heterogeneity and ``directionality'' risk.}
In decentralized settings, evaluators may differ not only in accuracy but also in \emph{directionality}---a metric can correlate negatively with the intended ground truth on some tasks or under some prompt styles.
This makes naive ensemble strategies risky: adding more signals can \emph{degrade} alignment if unreliable dimensions are not detected and calibrated.
Therefore, we seek a framework that (i) decomposes quality into interpretable dimensions, (ii) enables reliability auditing and task-wise calibration, and (iii) produces a composite score compatible with PoQ-style consensus and incentives.

\begin{table}[t]
  \centering
  \caption{Desired properties of quality signals for decentralized inference and PoQ integration.}
  \label{tab:desiderata}
  \begin{tabular}{@{}p{0.24\linewidth}p{0.72\linewidth}@{}}
    \toprule
    Property & Description \\
    \midrule
    Alignment & Correlates with human/ground-truth preference on target tasks; avoids reward hacking. \\
    Robustness & Stable under evaluator noise and resilient to adversarial evaluators or manipulations. \\
    Interpretability & Dimensions explain \emph{why} an output is scored highly/poorly, aiding audits and debugging. \\
    Cost awareness & Supports heterogeneous evaluation cost/latency and enables efficient sampling or tiered evaluation. \\
    Task adaptability & Allows task-specific weighting/calibration to mitigate metric mismatch across domains. \\
    Composability & Integrates as a drop-in module with PoQ aggregation, trust weighting, and reward mechanisms. \\
    \bottomrule
  \end{tabular}
\end{table}

\paragraph{Takeaway.}
PoQ provides a practical incentive backbone for decentralized inference, but it amplifies the importance of quality measurement.
This paper focuses on designing and validating a \emph{multi-dimensional} quality scoring framework that can be audited, calibrated, and then used as a PoQ-compatible quality signal.

\section{A Multi-Dimensional Quality Scoring Framework}
\label{sec:framework}

This section presents a modular framework for scoring LLM outputs with multiple interpretable dimensions and composing them into a single PoQ-compatible quality signal.
Our design goal is to keep each dimension \emph{auditable} and \emph{replaceable}, while enabling systematic reliability analysis and calibration before the signal is used for incentives.

\subsection{Design Goals and Principles}
\label{subsec:goals}

We design the framework around the desiderata in Table~\ref{tab:desiderata} and emphasize three additional principles.

\paragraph{Modularity.}
Each dimension is implemented as an independent scorer that maps a $(q,y)$ pair to a normalized scalar score.
Modularity allows dimensions to be added/removed without changing the rest of the pipeline, and facilitates ablations and failure analysis.

\paragraph{Auditability.}
Dimensions should be interpretable and tied to specific failure modes (e.g., degeneration, semantic drift, factual inconsistency).
This aligns with the broader observation that holistic evaluation requires breaking down model behavior into measurable components \citep{liang2023helm}.

\paragraph{Compatibility with decentralized incentives.}
The composite score should be suitable as a drop-in quality signal for PoQ-style aggregation and reward allocation \citep{tian2025costawarepoq,tian2026adaptiverobustpoq}.
In particular, it should support cost-aware evaluation (some dimensions are cheaper than others) and be robust to evaluator heterogeneity.

\begin{figure}[t]
  \centering
  \resizebox{1.0\linewidth}{!}{%
  \begin{tikzpicture}[
    box/.style={
      draw, rounded corners, align=center,
      inner sep=5pt, minimum width=18mm, minimum height=12mm
    },
    arrow/.style={-Latex, line width=0.7pt},
    darrow/.style={-Latex, dashed, line width=0.6pt}
  ]

    \node[box] (comp)  at (0.0, 0.0) {Composite\\Quality Score $\hat{s}(q,y)$};

    \node[box] (sem)   at (-3.8, 2.2) {Semantic\\Quality};
    \node[box] (align) at ( 3.8, 2.2) {Query-Output\\Alignment};

    \node[box] (struct) at (-5.6, 0.0) {Structure\\Quality};

    \node[box] (prior) at (-3.8,-2.2) {Priors\\(model, cost)};
    \node[box] (agree) at ( 3.8,-2.2) {Agreement /\\Uncertainty};

    \node[box] (qy)    at (-9.2, 0.0) {$q$ (query), $y$ (output)};
    \node[box] (poq)   at ( 9.0, 0.0) {PoQ Aggregation\\and Rewards};

    \node[draw, rounded corners, inner sep=10pt,
          fit=(sem)(align)(struct)(prior)(agree)(comp),
          label=above:{Dimension Modules}] (dims) {};

    \draw[arrow]  (struct) -- (comp);
    \draw[arrow]  (sem)   -- (comp);
    \draw[arrow]  (align) -- (comp);
    \draw[arrow]  (prior) -- (comp);
    \draw[arrow]  (agree) -- (comp);

    \draw[arrow]  (comp) -- (poq);

    \draw[arrow]  (qy) -- (struct);
    \draw[darrow] (qy) -- (sem);
    \draw[darrow] (qy) -- (align);
    \draw[darrow] (qy) -- (prior);
    \draw[darrow] (qy) -- (agree);

  \end{tikzpicture}%
  }
  \caption{Modular architecture of multi-dimensional quality scoring. Each dimension module produces a normalized score; the composite score $\hat{s}(q,y)$ is then used as a PoQ-compatible quality signal for aggregation and incentives.}
  \label{fig:dimension-modules}
\end{figure}
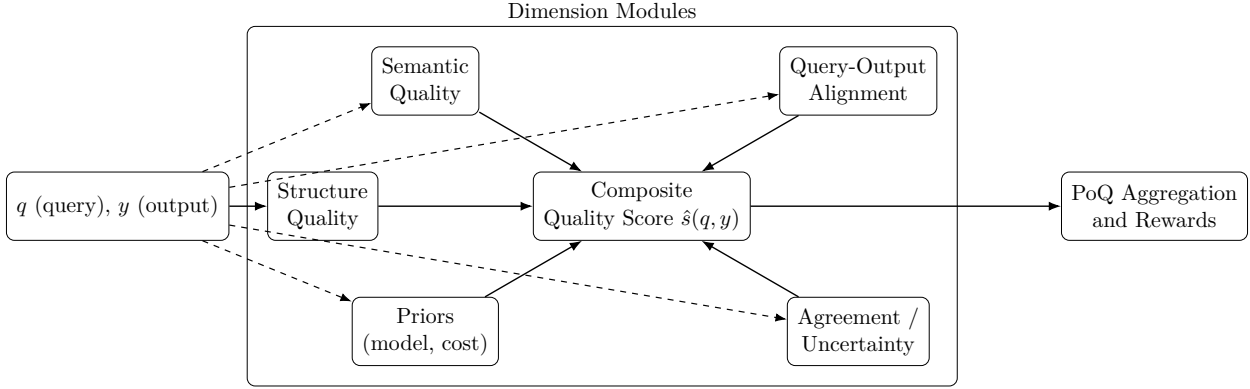

\subsection{Quality Dimensions}
\label{subsec:dimensions}

We organize dimensions into five families. Let $z_k(q,y)\in[0,1]$ denote the normalized score for dimension $k$.

\paragraph{(1) Priors.}
Priors provide weak but cheap signals that are useful for cold-start and for regularizing noisy dimensions.
We use (i) a model prior based on preference-derived rankings (e.g., Elo/TrueSkill-style ratings) \citep{herbrich2007trueskill,chiang2024chatbotarena}, and (ii) a cost-efficiency prior capturing a model's quality-per-cost tendency under a fixed serving budget.

\paragraph{(2) Structural quality.}
Structural features detect formatting violations, excessive repetition, unusually short/long outputs, and degeneration patterns.
These heuristics are lightweight and can filter clear failures before applying more expensive evaluators.

\paragraph{(3) Semantic quality.}
Semantic similarity aims to measure whether the output preserves meaning relative to a reference or target (task-dependent).
We instantiate this dimension with sentence embedding methods such as Sentence-BERT and contrastive variants \citep{reimers-gurevych-2019-sentence,gao-etal-2021-simcse}, as well as embedding-based generation metrics \citep{zhang2020bertscore,zhao-etal-2019-moverscore}.

\paragraph{(4) Query-output alignment.}
Alignment measures instruction-following and entailment-like consistency between the query and output.
NLI-style and learned evaluators are commonly used for consistency detection in summarization and instruction compliance \citep{laban-etal-2022-summac,kryscinski-etal-2020-evaluating}.
We treat alignment as \emph{task-sensitive} and subject to calibration.

\paragraph{(5) Agreement / uncertainty.}
When multiple evaluators are available, disagreement can be used as a proxy for uncertainty.
However, disagreement can be misleading under evaluator heterogeneity or when some evaluators exhibit reversed directionality.
Thus, this dimension must be validated and possibly down-weighted, especially in adversarial settings \citep{blanchard2017krum,pmlr-v80-yin18a,mhamdi2018hidden}.

\begin{table}[t]
  \centering
  \caption{Dimension definitions and typical instantiations used in our framework. Weights are default values used as a starting point; later sections perform ablations and task-wise calibration.}
  \label{tab:dimensions}
  \begin{tabular}{@{}p{0.22\linewidth}p{0.58\linewidth}p{0.14\linewidth}@{}}
    \toprule
    Dimension & Instantiation (examples) & Default weight \\
    \midrule
    Model prior & Preference-derived model ranking (Elo/TrueSkill style) \citep{herbrich2007trueskill,chiang2024chatbotarena} & 0.15 \\
    Cost-efficiency prior & Quality-per-cost tendency under a serving budget (offline benchmark) & 0.10 \\
    Structure quality & Length/format checks, repetition/degeneration signals & 0.20 \\
    Semantic quality & Sentence embeddings / learned semantic metrics \citep{reimers-gurevych-2019-sentence,gao-etal-2021-simcse,zhang2020bertscore} & 0.25 \\
    Query-output alignment & NLI/consistency style evaluators \citep{laban-etal-2022-summac,kryscinski-etal-2020-evaluating} & 0.15 \\
    Agreement / uncertainty & Cross-evaluator dispersion / uncertainty proxy \citep{pmlr-v80-yin18a,mhamdi2018hidden} & 0.15 \\
    \bottomrule
  \end{tabular}
\end{table}

\section{Experimental Methodology}
\label{sec:method}

We evaluate the proposed multi-dimensional scoring framework from two perspectives:
(1) \emph{quality-signal alignment}, i.e., how well each dimension and the composite correlate with reference quality signals; and
(2) \emph{mechanism-level impact}, i.e., how the composite behaves when used as the quality signal in PoQ-style aggregation and reward allocation.
Our experimental protocol follows the same decentralized inference and evaluator setting used in our prior PoQ studies to ensure comparability \citep{tian2025costawarepoq,tian2026adaptiverobustpoq}.

\subsection{Tasks and Datasets}
\label{subsec:data}

We consider two representative task families:
(i) question answering (QA), where correctness is sensitive to instruction compliance and exactness, and
(ii) summarization, where semantic coverage and factual consistency are central.
This mix allows us to stress-test the task dependence of different quality dimensions, particularly alignment- and agreement-based signals.

\subsection{Inference Models and Evaluators}
\label{subsec:models}

\paragraph{Inference model pool.}
We evaluate multiple inference models to capture heterogeneity typical in decentralized settings.
This includes models with different quality and cost profiles, which motivates the use of priors and cost-aware scoring \citep{tian2025costawarepoq}.

\paragraph{Evaluator pool.}
We use a set of automatic evaluators spanning semantic similarity metrics, learned text-quality metrics, and NLI-style or consistency-oriented models.
This reflects a realistic PoQ deployment where evaluators are diverse and may disagree due to bias or task mismatch \citep{sellam-etal-2020-bleurt,rei-etal-2020-comet,laban-etal-2022-summac}.

\begin{table}[t]
  \centering
  \caption{Experimental setup overview. Fill in concrete counts/models to match your run logs.}
  \label{tab:setup}
  \begin{tabular}{@{}p{0.28\linewidth}p{0.66\linewidth}@{}}
    \toprule
    Component & Configuration \\
    \midrule
    Tasks & QA; Summarization \\
    Inference models & Multiple heterogeneous LLMs (quality/cost diversity) \\
    Evaluators & Semantic metrics; learned metrics; NLI/consistency evaluators \\
    Dimensions ($K$) & Priors; structure; semantics; alignment; agreement/uncertainty \\
    Composite form & Weighted sum with normalization and clipping (Eq.~\ref{eq:composite}) \\
    PoQ simulation & Cost-aware sampling; robust aggregation; trust weighting \citep{tian2025costawarepoq,tian2026adaptiverobustpoq} \\
    \bottomrule
  \end{tabular}
\end{table}

\subsection{Ground Truth and Reference Signals}
\label{subsec:gt}

\paragraph{Reference quality signals.}
To measure alignment of dimensions and composite scoring, we compare against reference signals representing target quality.
These can include human annotations (when available) and/or strong judge signals (e.g., preference-based leaderboards or LLM-as-a-judge protocols).
We treat such signals as \emph{reference} rather than absolute ground truth, consistent with the literature on holistic evaluation and preference-based benchmarking \citep{liang2023helm,zheng2023judging,chiang2024chatbotarena}.

\paragraph{Why multiple references.}
No single metric perfectly captures quality across tasks.
For example, overlap-based metrics are known to be brittle for summarization \citep{lin-2004-rouge}, while learned metrics can encode biases and may not transfer \citep{sellam-etal-2020-bleurt,rei-etal-2020-comet}.
Therefore, our analysis emphasizes consistency across references and tasks.

\subsection{Metrics and Analysis Protocol}
\label{subsec:metrics}

\paragraph{Correlation analysis.}
For each dimension score $z_k(q,y)$ and the composite score $\hat{s}(q,y)$, we compute correlation with reference signals using Pearson and Spearman correlations.
We report both because Pearson measures linear association while Spearman is rank-based and more robust to monotonic transformations.

\paragraph{Ablations and weight sensitivity.}
We conduct ablations by (i) removing a dimension, and (ii) varying weights $\{w_k\}$ to quantify how sensitive alignment is to dimension selection and calibration.
This is essential because adding dimensions can reduce alignment if a dimension is task-mismatched or directionally inverted.

\paragraph{Task-wise decomposition.}
We evaluate correlations separately on QA and summarization subsets to expose task dependence.
This is particularly important for alignment- and agreement-based dimensions, which may behave differently across tasks \citep{kryscinski-etal-2020-evaluating,scialom-etal-2021-questeval,laban-etal-2022-summac}.

\paragraph{PoQ integration evaluation.}
To connect quality-signal design to decentralized incentives, we plug the composite score into PoQ aggregation and reward mechanisms.
We reuse the cost-aware and robust PoQ components developed in our prior work \citep{tian2025costawarepoq,tian2026adaptiverobustpoq}, and evaluate:
(i) reward ranking consistency with reference quality, and
(ii) robustness under evaluator heterogeneity and adversarial conditions.
The robust aggregation perspective is conceptually related to Byzantine-resilient estimation and robust learning \citep{castro1999pbft,blanchard2017krum,pmlr-v80-yin18a,mhamdi2018hidden}.

\begin{table}[t]
  \centering
  \caption{Example cost/latency tiers for dimensions and evaluators. The purpose is to support cost-aware evaluation and PoQ sampling \citep{tian2025costawarepoq}. Replace ``Low/Med/High'' with measured values if available.}
  \label{tab:cost-tiers}
  \begin{tabular}{@{}p{0.30\linewidth}p{0.34\linewidth}p{0.28\linewidth}@{}}
    \toprule
    Module & Typical computation & Cost tier \\
    \midrule
    Priors & Lookup / offline benchmark & Low \\
    Structure quality & Heuristics over text & Low \\
    Semantic quality & Embedding similarity \citep{reimers-gurevych-2019-sentence,gao-etal-2021-simcse} & Medium \\
    Query-output alignment & NLI/consistency evaluator \citep{laban-etal-2022-summac} & Medium--High \\
    Agreement/uncertainty & Cross-evaluator dispersion & Depends on evaluator set \\
    \bottomrule
  \end{tabular}
\end{table}

\paragraph{Reproducibility.}
All scores are computed on logged $(q,y)$ pairs.
Composite scoring is implemented as a deterministic post-processing layer given dimension outputs, enabling rapid experimentation with different weightings and ablations without rerunning inference.

\section{Results: Dimension Reliability and Composite Alignment}
\label{sec:results}

We now present empirical results on the reliability of individual dimensions and the composite quality score.
Our analysis focuses on alignment with the reference quality signal (``GT'') and emphasizes two questions:
(i) which dimensions are consistently informative, and
(ii) when does a composite score improve or degrade alignment relative to strong single evaluators and consensus baselines.

\subsection{Correlation Landscape Across Evaluators, Consensus, and Composite}
\label{subsec:single-dim}

\begin{figure}[t]
  \centering
  \includegraphics[width=0.9\linewidth]{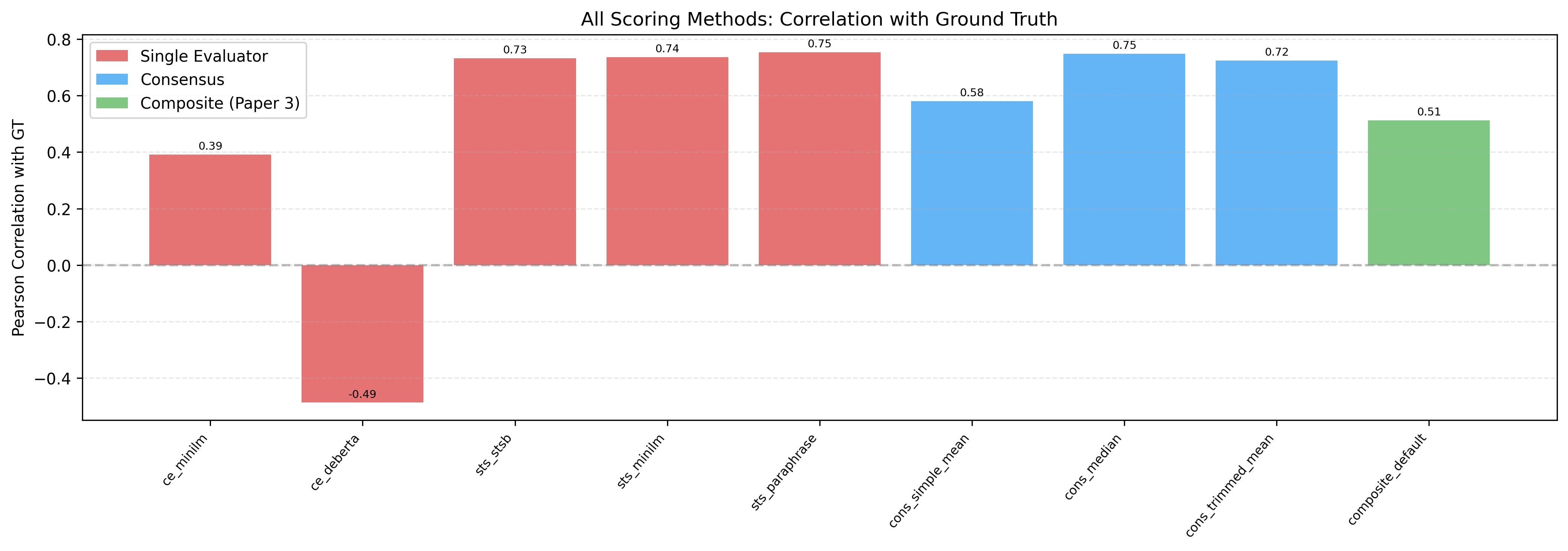}
  \caption{Unified correlation summary across individual evaluators, consensus methods, and the default composite score.}
  \label{fig:correlation-unified}
\end{figure}

Figure~\ref{fig:correlation-unified} summarizes the overall correlation landscape across individual evaluators, consensus methods, and the default composite.
Table~\ref{tab:overall-corr} reports the key correlation values (Pearson and Spearman) used throughout this section.

\begin{table}[t]
  \centering
  \caption{Overall correlations with the reference signal (GT) on 2000 samples.}
  \label{tab:overall-corr}
  \begin{tabular}{@{}lcc@{}}
    \toprule
    Method / Dimension & Pearson $\uparrow$ & Spearman $\uparrow$ \\
    \midrule
    Composite (default weights) & 0.513 & 0.682 \\
    Best single evaluator (sts\_paraphrase) & 0.754 & 0.806 \\
    Best consensus baseline (median over evaluators) & 0.749 & 0.801 \\
    \midrule
    Semantic quality (dimension) & 0.733 & 0.776 \\
    Structure quality (dimension) & 0.466 & 0.601 \\
    Query-output alignment (dimension) & -0.437 & -0.238 \\
    Agreement / uncertainty (dimension) & -0.384 & 0.008 \\
    \bottomrule
  \end{tabular}
\end{table}

Two immediate observations follow.
First, the default composite score does \emph{not} match the strongest single learned semantic evaluator or the strongest consensus baseline (median), despite combining more signals.
Second, two intuitive dimensions---query-output alignment and agreement/uncertainty---exhibit negative Pearson correlation overall, indicating that naive inclusion can \emph{harm} alignment.

\begin{figure}[t]
  \centering
  \includegraphics[width=0.6\linewidth]{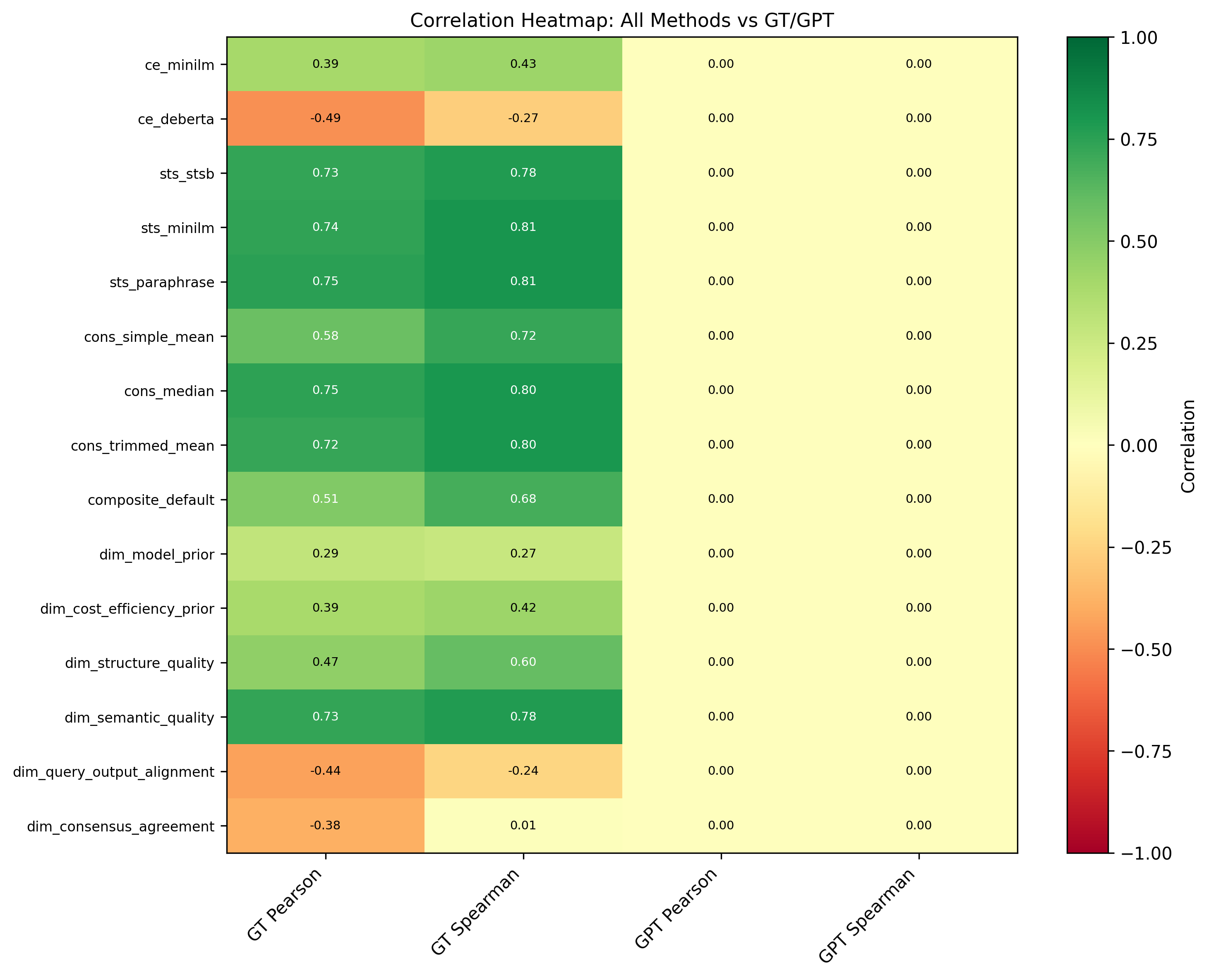}
  \caption{Correlation heatmap (GT) across evaluators, consensus baselines, composite, and dimensions.}
  \label{fig:correlation-heatmap}
\end{figure}

For a more granular view across methods and dimensions, Figure~\ref{fig:correlation-heatmap} provides a correlation heatmap.

\subsection{Single-Dimension Reliability and Task Dependence}
\label{subsec:dim-reliability}

\begin{figure}[t]
  \centering
  \includegraphics[width=0.8\linewidth]{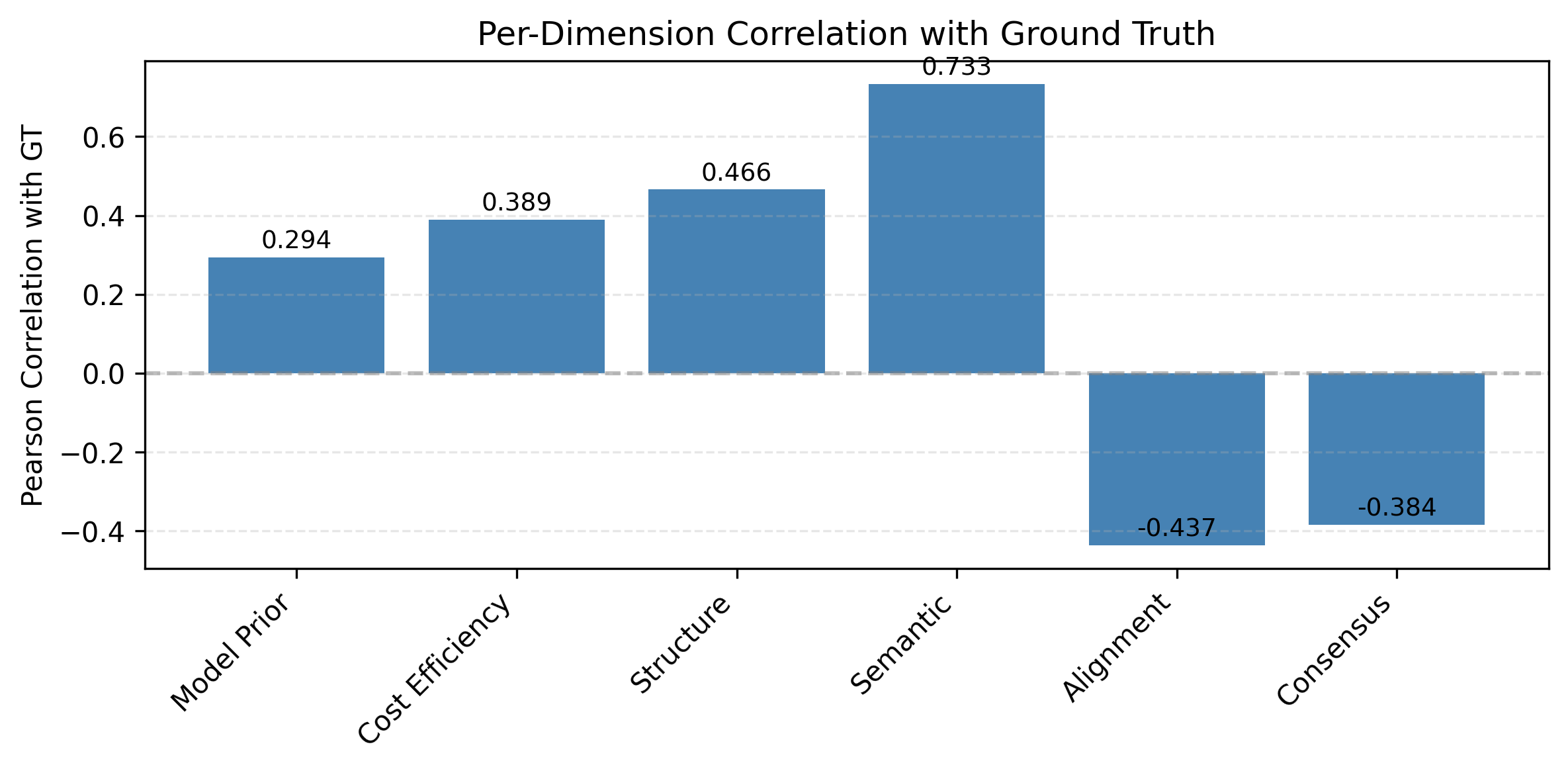}
  \caption{Per-dimension correlation with GT. Semantic quality is strongly aligned overall, while alignment and agreement dimensions can be negatively correlated without calibration.}
  \label{fig:dimension-gt-corr}
\end{figure}

Figure~\ref{fig:dimension-gt-corr} reports GT correlations per dimension, highlighting a strong semantic dimension and weaker priors/structure signals.
Importantly, alignment and agreement dimensions can be unreliable without calibration.

To diagnose whether negative correlations are uniform across tasks, Table~\ref{tab:taskwise} decomposes correlations by QA vs summarization.
Alignment and agreement are strongly \emph{negative} on QA but become weakly \emph{positive} on summarization, revealing sharp task dependence.
This motivates task-aware calibration rather than assuming a dimension is universally beneficial.

\begin{table}[t]
  \centering
  \caption{Task-wise correlations (Pearson / Spearman) with GT. ``Composite (calibrated)'' removes the alignment and agreement dimensions and re-normalizes the remaining weights.}
  \label{tab:taskwise}
  \begin{tabular}{@{}lcc@{}}
    \toprule
    Signal (vs.\ GT) & QA & Summarization \\
    \midrule
    Semantic quality & 0.863 / 0.796 & 0.433 / 0.357 \\
    Query-output alignment & -0.571 / -0.493 & 0.102 / 0.058 \\
    Agreement / uncertainty & -0.553 / -0.401 & 0.188 / 0.239 \\
    Composite (default) & 0.742 / 0.720 & 0.443 / 0.408 \\
    Composite (calibrated) & 0.893 / 0.795 & 0.438 / 0.408 \\
    \bottomrule
  \end{tabular}
\end{table}

\begin{figure}[t]
  \centering
  \includegraphics[width=0.8\linewidth]{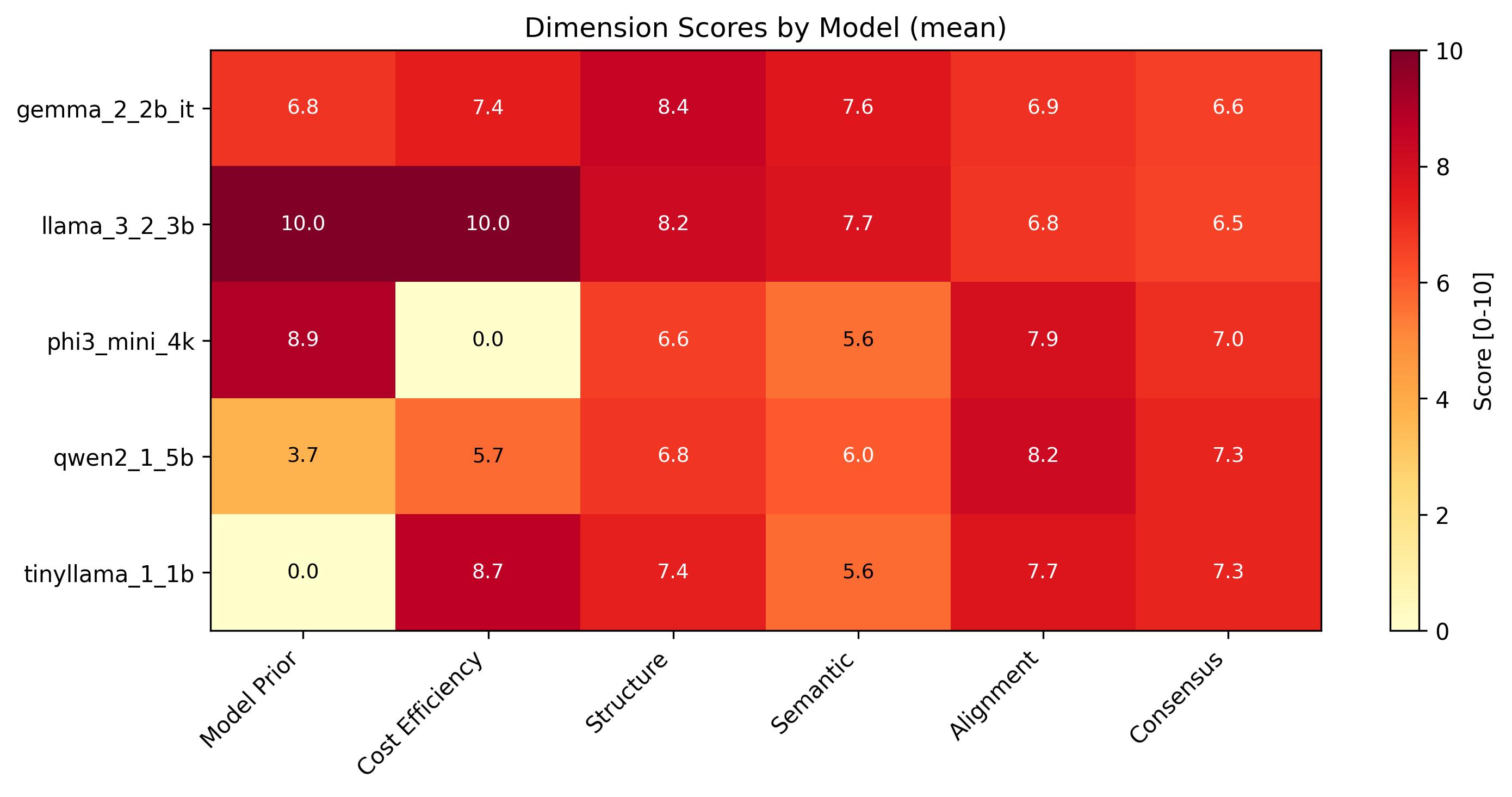}
  \caption{Dimension heatmap visualization for diagnosing dimension behavior across settings (e.g., models/tasks).}
  \label{fig:dimension-heatmap}
\end{figure}

\begin{figure}[t]
  \centering
  \includegraphics[width=0.6\linewidth]{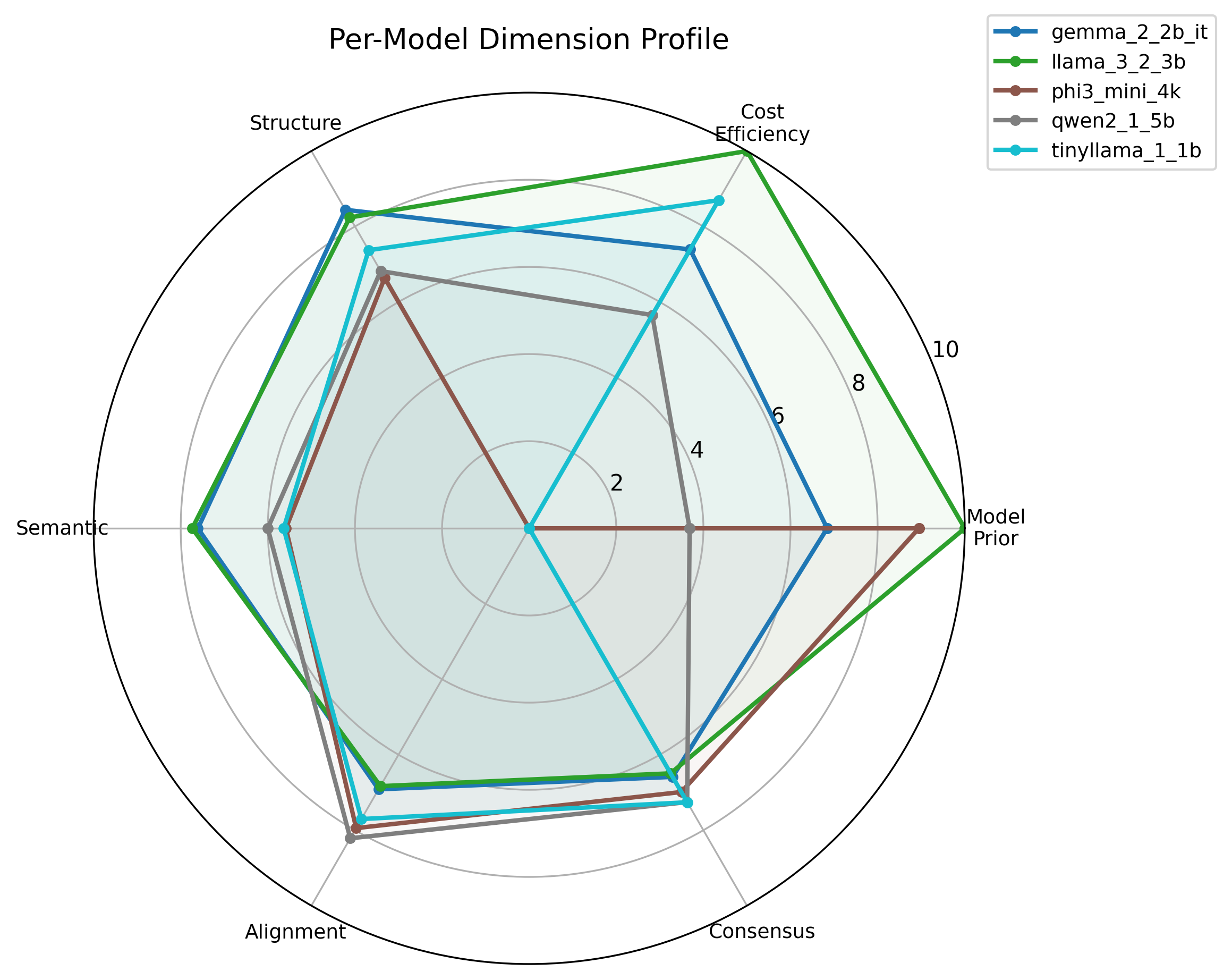}
  \caption{Radar chart view of multi-dimensional scores for qualitative comparison across outputs or model groups.}
  \label{fig:dimension-radar}
\end{figure}

Figures~\ref{fig:dimension-heatmap} and \ref{fig:dimension-radar} provide additional views of dimension behavior and cross-dimension structure.

\subsection{Composite vs.\ Single Evaluators}
\label{subsec:composite-vs-single}

\begin{figure}[t]
  \centering
  \includegraphics[width=0.5\linewidth]{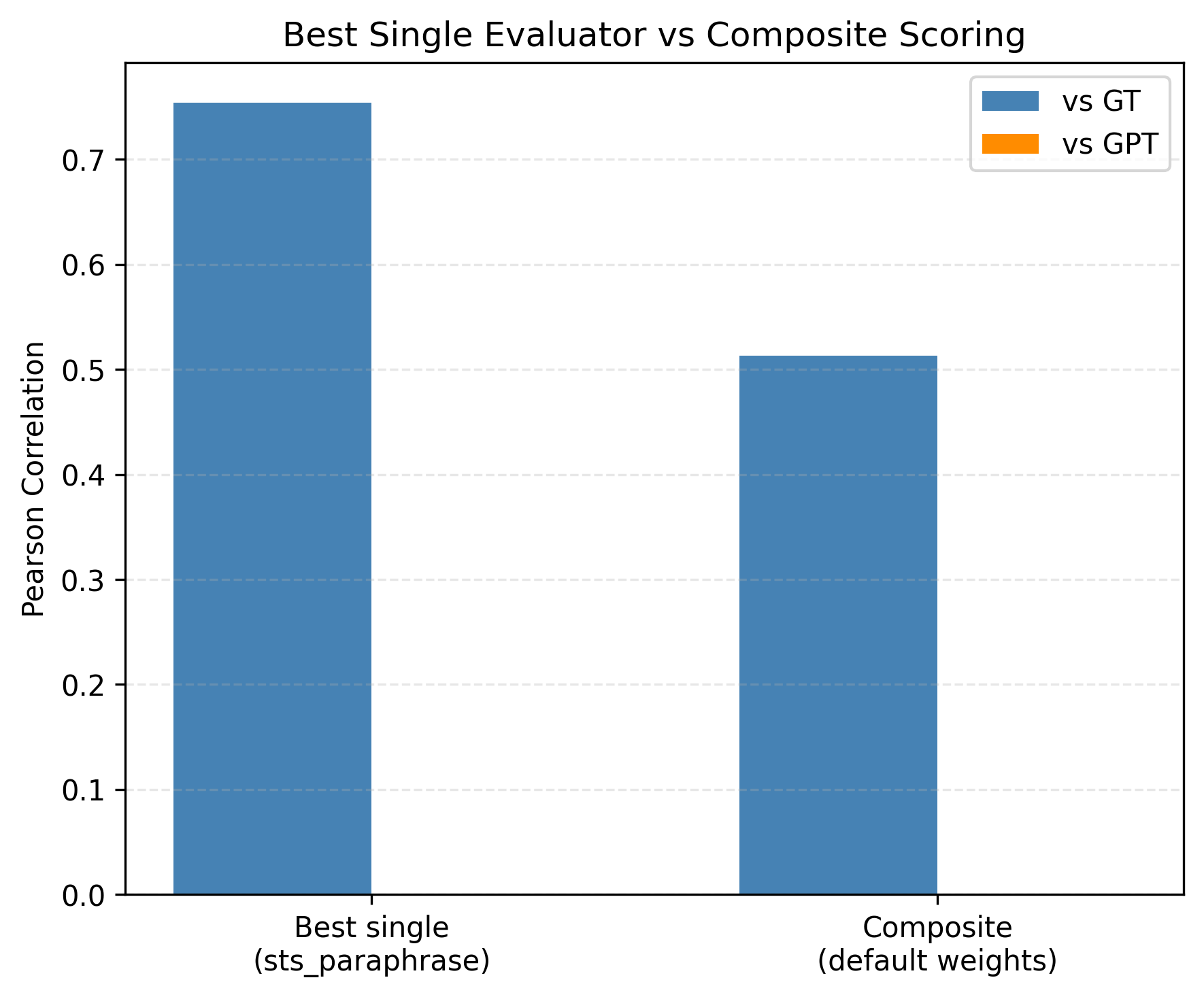}
  \caption{Composite score versus the best single evaluator baseline. Despite using multiple signals, the default composite can underperform the strongest semantic evaluator without calibration.}
  \label{fig:composite-vs-best}
\end{figure}

\begin{figure}[t]
  \centering
  \includegraphics[width=0.7\linewidth]{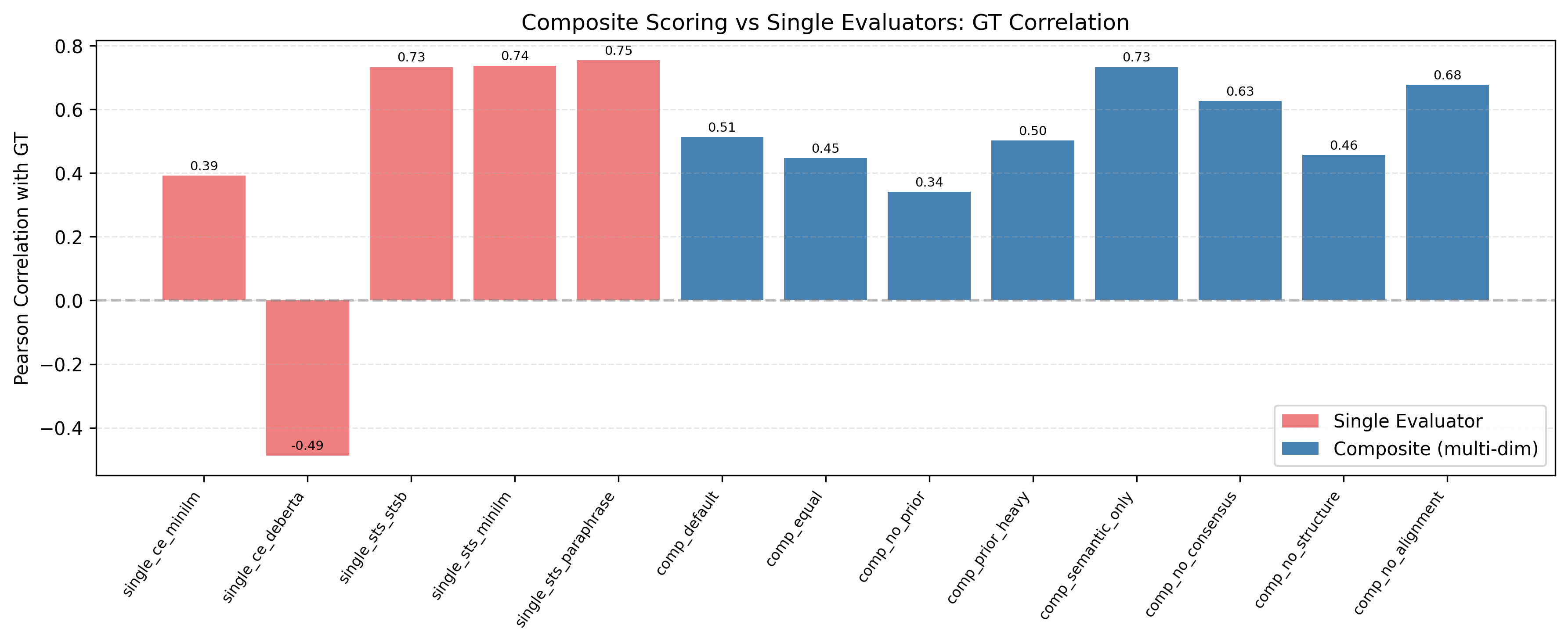}
  \caption{Composite and single-evaluator correlations against GT. This view highlights when composites help and when they degrade alignment relative to individual evaluators.}
  \label{fig:composite-vs-single-gt}
\end{figure}

We compare the composite score to strong single evaluators and consensus baselines.
Figure~\ref{fig:composite-vs-best} contrasts the composite against the strongest single evaluator, while Figure~\ref{fig:composite-vs-single-gt} shows broader comparisons against GT.

The key takeaway is that multi-dimensional scoring is not ``free'':
if one or more dimensions are directionally inverted or task-mismatched, the composite can be dragged below strong single baselines.
This motivates explicit reliability auditing and calibration.

\subsection{Ablations and Calibration}
\label{subsec:ablations}

\begin{figure}[t]
  \centering
  \includegraphics[width=0.7\linewidth]{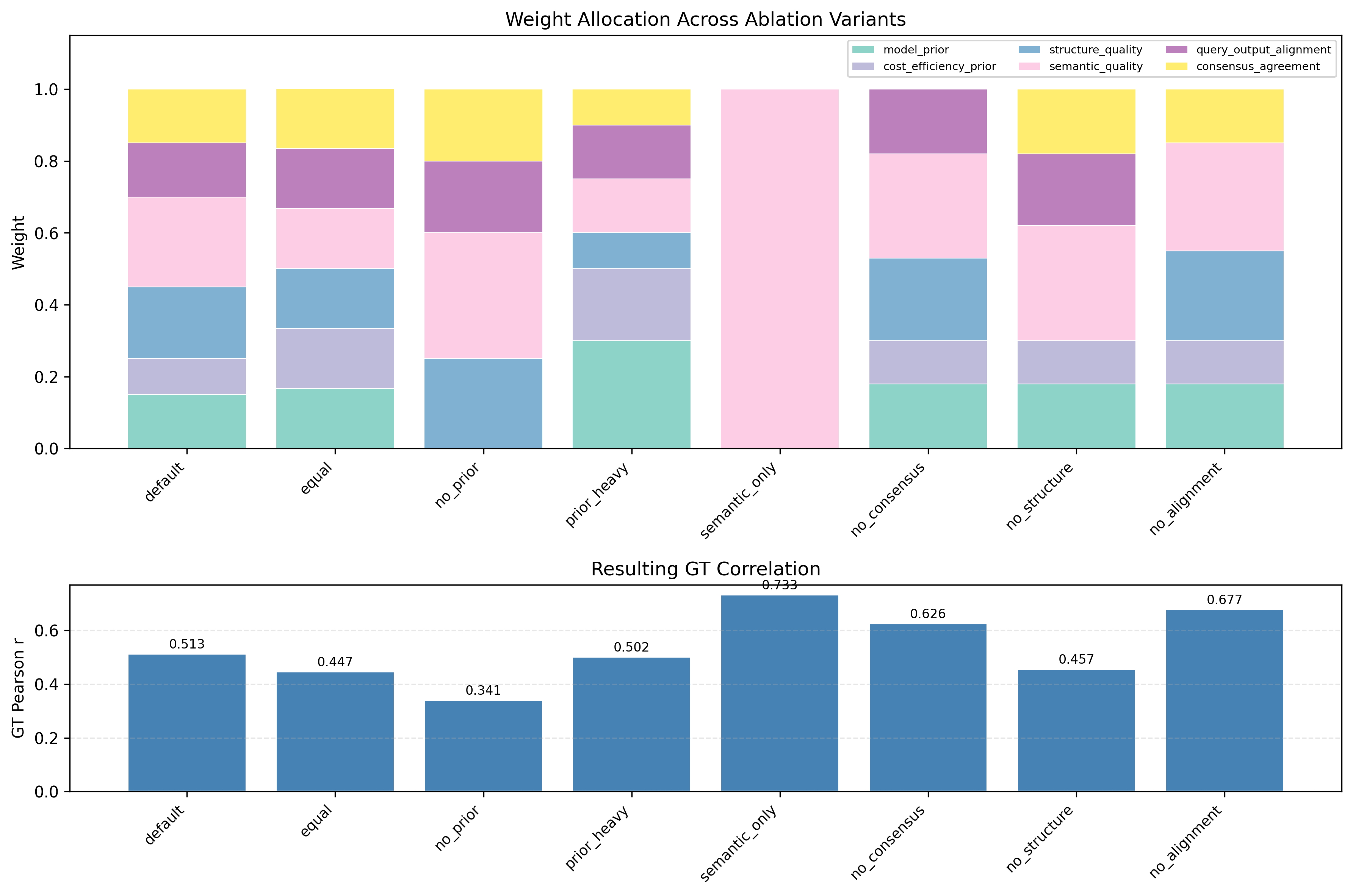}
  \caption{Weight ablation and dimension removal results. Removing unreliable dimensions can substantially improve alignment.}
  \label{fig:weight-ablation}
\end{figure}

Figure~\ref{fig:weight-ablation} studies how performance changes under weight variants and dimension removals.
Table~\ref{tab:ablation} lists representative ablation variants and their GT correlations.

\begin{table}[t]
  \centering
  \caption{Ablation variants and correlations with GT. ``Calibrated (remove alignment+agreement)'' is computed by removing those dimensions and re-normalizing remaining weights.}
  \label{tab:ablation}
  \begin{tabular}{@{}lcc@{}}
    \toprule
    Variant & Pearson $\uparrow$ & Spearman $\uparrow$ \\
    \midrule
    Default & 0.513 & 0.682 \\
    Equal weights & 0.447 & 0.643 \\
    No priors & 0.341 & 0.543 \\
    Prior-heavy & 0.502 & 0.613 \\
    Semantic only & 0.733 & 0.776 \\
    No structure & 0.457 & 0.619 \\
    No alignment & 0.677 & 0.778 \\
    No agreement & 0.626 & 0.727 \\
    \midrule
    Calibrated (remove alignment+agreement) & 0.760 & 0.800 \\
    \bottomrule
  \end{tabular}
\end{table}

Overall, the ablation results reinforce a central message:
\emph{multi-dimensional} scoring becomes valuable when it is paired with \emph{dimension reliability auditing} and \emph{calibration}.
In our setting, removing the unreliable alignment and agreement dimensions yields a calibrated composite that slightly surpasses the strongest single evaluator and median consensus baseline in overall Pearson correlation (Table~\ref{tab:ablation}).

\paragraph{Interpreting failures.}
Negative alignment correlation is consistent with the observation that NLI-style or instruction-following signals can be task-sensitive and may not transfer as intended, especially when the reference signal rewards a different notion of correctness.
Similarly, agreement-based uncertainty proxies can become misleading under evaluator heterogeneity or when some evaluators exhibit reversed directionality, echoing robustness concerns in Byzantine settings \citep{blanchard2017krum,pmlr-v80-yin18a,mhamdi2018hidden}.

\section{PoQ Integration as an Application Study}
\label{sec:poq-integration}

This section treats the proposed composite score as a \emph{drop-in} quality signal for Proof of Quality (PoQ) and evaluates its mechanism-level impact.
The goal is not to redesign PoQ itself (covered in our prior work), but to show how quality-signal design influences reward ranking, robustness, and adversarial resilience when deployed within PoQ-style aggregation and incentives \citep{tian2025costawarepoq,tian2026adaptiverobustpoq}.

\subsection{Using Composite Scores as PoQ Quality Signals}
\label{subsec:poq-quality}

\paragraph{Integration interface.}
From PoQ's perspective, replacing a single evaluator score with a composite quality score $\hat{s}(q,y)$ requires no change to the core protocol.
PoQ continues to (i) sample evaluators subject to cost constraints, (ii) aggregate scores into a consensus estimate, and (iii) allocate rewards based on consensus quality \citep{tian2025costawarepoq}.
Therefore, the composite score can be deployed as a modular quality layer without disrupting the incentive mechanism.

\begin{figure}[t]
  \centering
  \includegraphics[width=1.0\linewidth]{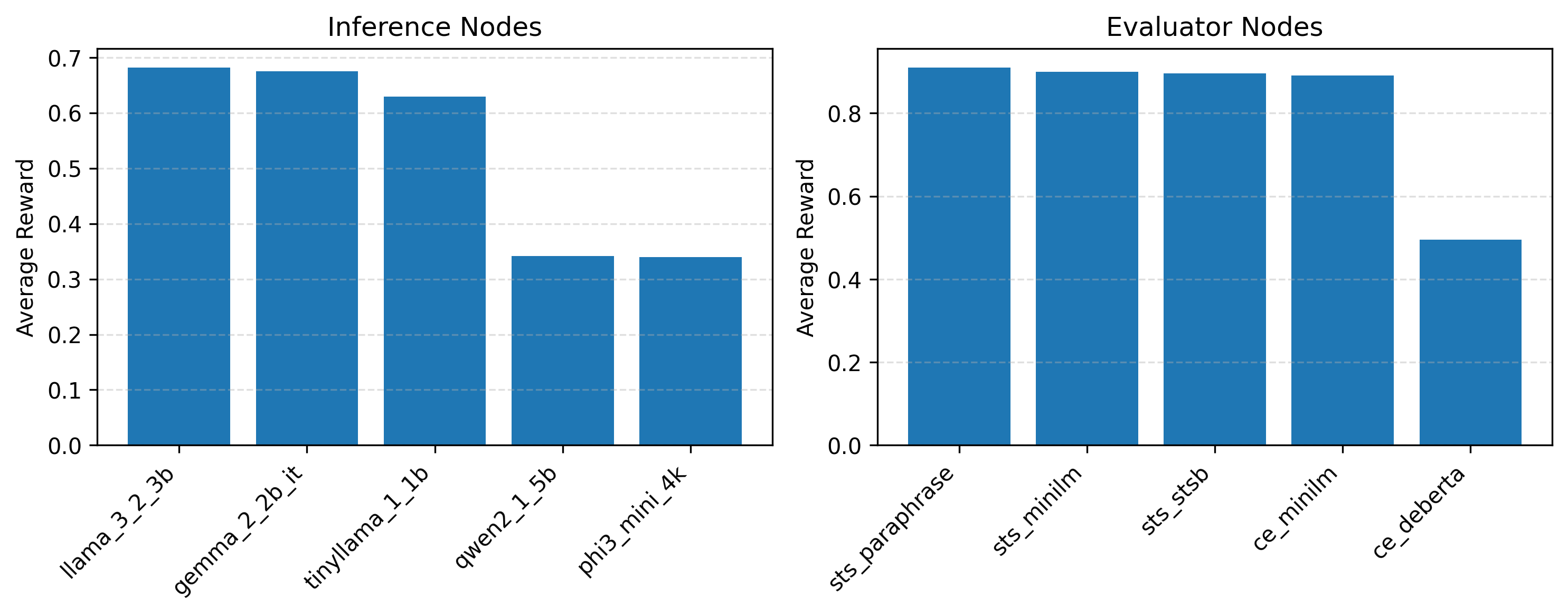}
  \caption{PoQ reward outcomes under baseline quality signals. The reward landscape reflects the alignment properties of the quality signal used for consensus and incentives.}
  \label{fig:baseline-rewards}
\end{figure}

\paragraph{Reward ranking and incentive implications.}
Figure~\ref{fig:baseline-rewards} shows how reward outcomes differ across baseline quality signals.
A key takeaway is that PoQ inherits the strengths and weaknesses of the underlying quality signal: if the signal is misaligned, the reward landscape can systematically favor suboptimal behaviors.
This reinforces the motivation for the reliability auditing and calibration of Section~\ref{sec:results}.

\subsection{Interaction with Robust Aggregation and Trust Weighting}
\label{subsec:poq-robust}

\paragraph{Robust aggregation complements composite scoring.}
Robust aggregation mechanisms (e.g., median, trimmed mean) are designed to reduce the influence of outliers or malicious evaluators.
These methods are conceptually related to Byzantine-resilient estimation and robust learning, where naive averaging can be fragile \citep{castro1999pbft,blanchard2017krum,pmlr-v80-yin18a,mhamdi2018hidden}.
Our prior work shows that adaptive trust weighting further improves robustness by discounting evaluators that behave inconsistently over time \citep{tian2026adaptiverobustpoq}.

\begin{figure}[t]
  \centering
  \includegraphics[width=0.6\linewidth]{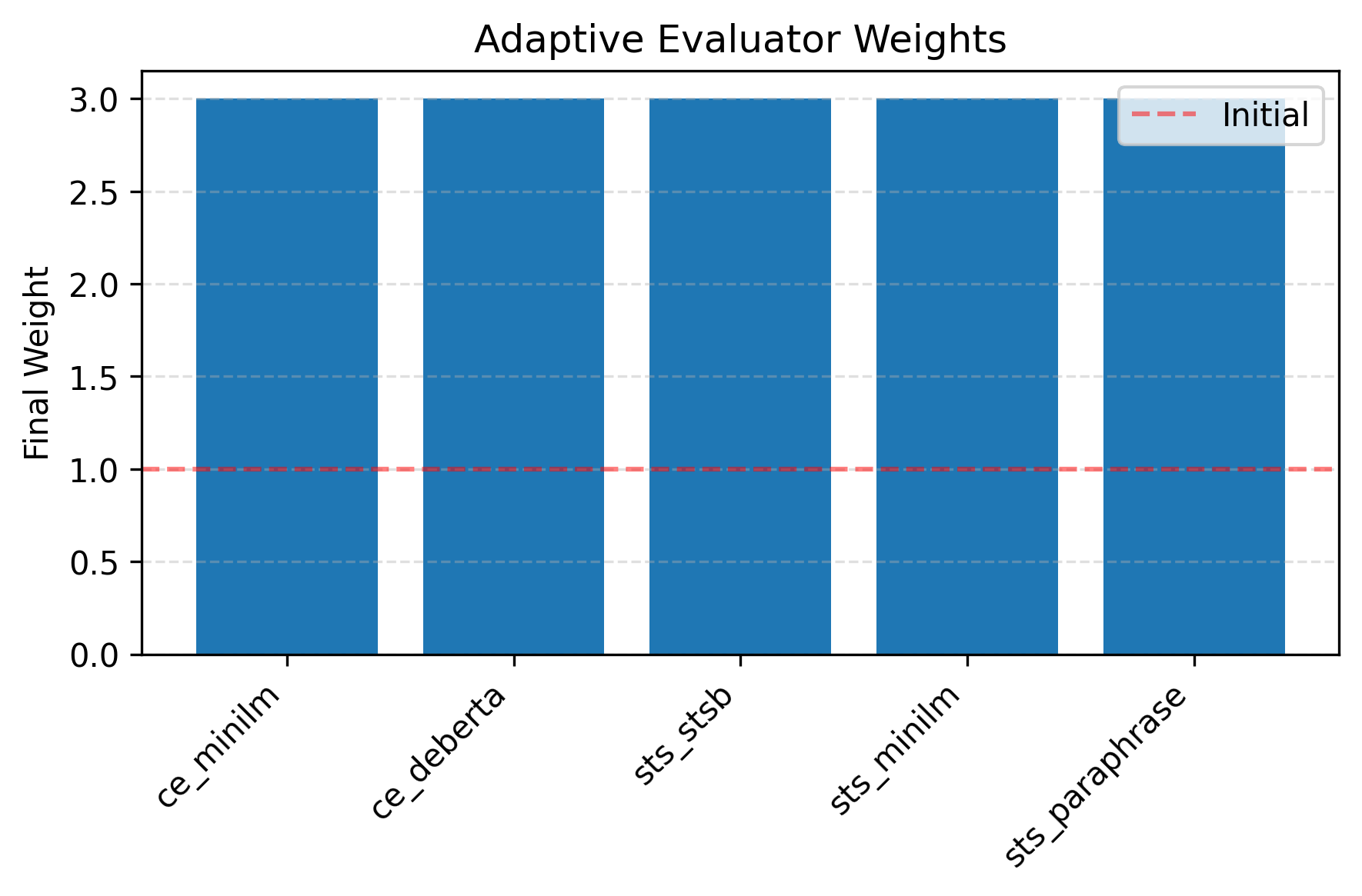}
  \caption{Adaptive trust weighting behavior in PoQ. Trust-weight updates help suppress unreliable evaluators and complement multi-dimensional scoring.}
  \label{fig:adaptive-weights}
\end{figure}

\paragraph{Composite scoring with adaptive trust.}
Multi-dimensional scoring and adaptive trust weighting address different failure modes:
the composite focuses on \emph{measurement design} (what is being measured), while trust weighting focuses on \emph{source reliability} (who is measuring).
Figure~\ref{fig:adaptive-weights} illustrates how adaptive weights evolve in response to evaluator behavior, providing a mechanism to down-weight unreliable evaluators even when the quality signal is noisy.

\begin{table}[t]
  \centering
  \caption{PoQ integration study configurations. We evaluate the composite score under a grid of attack types, attack ratios, and defense mechanisms.}
  \label{tab:poq-config}
  \begin{tabular}{@{}p{0.28\linewidth}p{0.66\linewidth}@{}}
    \toprule
    Component & Settings \\
    \midrule
    Quality signal & Single evaluator; consensus baseline; composite (default / calibrated) \\
    Aggregation defenses & Mean; Median; Trimmed mean; Adaptive trust weighting \citep{tian2026adaptiverobustpoq} \\
    Attack types & Multiple malicious evaluator strategies (see Fig.~\ref{fig:defense-summary}) \\
    Attack ratios & Multiple attacker fractions (swept) \\
    Evaluation budget & Cost-aware sampling / fixed budget variants \citep{tian2025costawarepoq} \\
    Simulation & Monte Carlo rounds per configuration \\
    \bottomrule
  \end{tabular}
\end{table}

\subsection{Adversarial Considerations}
\label{subsec:poq-adversarial}

\paragraph{Threat model.}
Decentralized inference networks must assume that some evaluator nodes can be malicious, colluding, or strategically biased.
Attacks may target the consensus score directly (score inflation/deflation), or exploit evaluator mismatch to distort reward allocation.
This is closely related to Byzantine settings studied in distributed systems and robust learning \citep{castro1999pbft,blanchard2017krum,pmlr-v80-yin18a,mhamdi2018hidden}.

\begin{figure}[!htbp]
  \centering
  \includegraphics[width=0.6\linewidth]{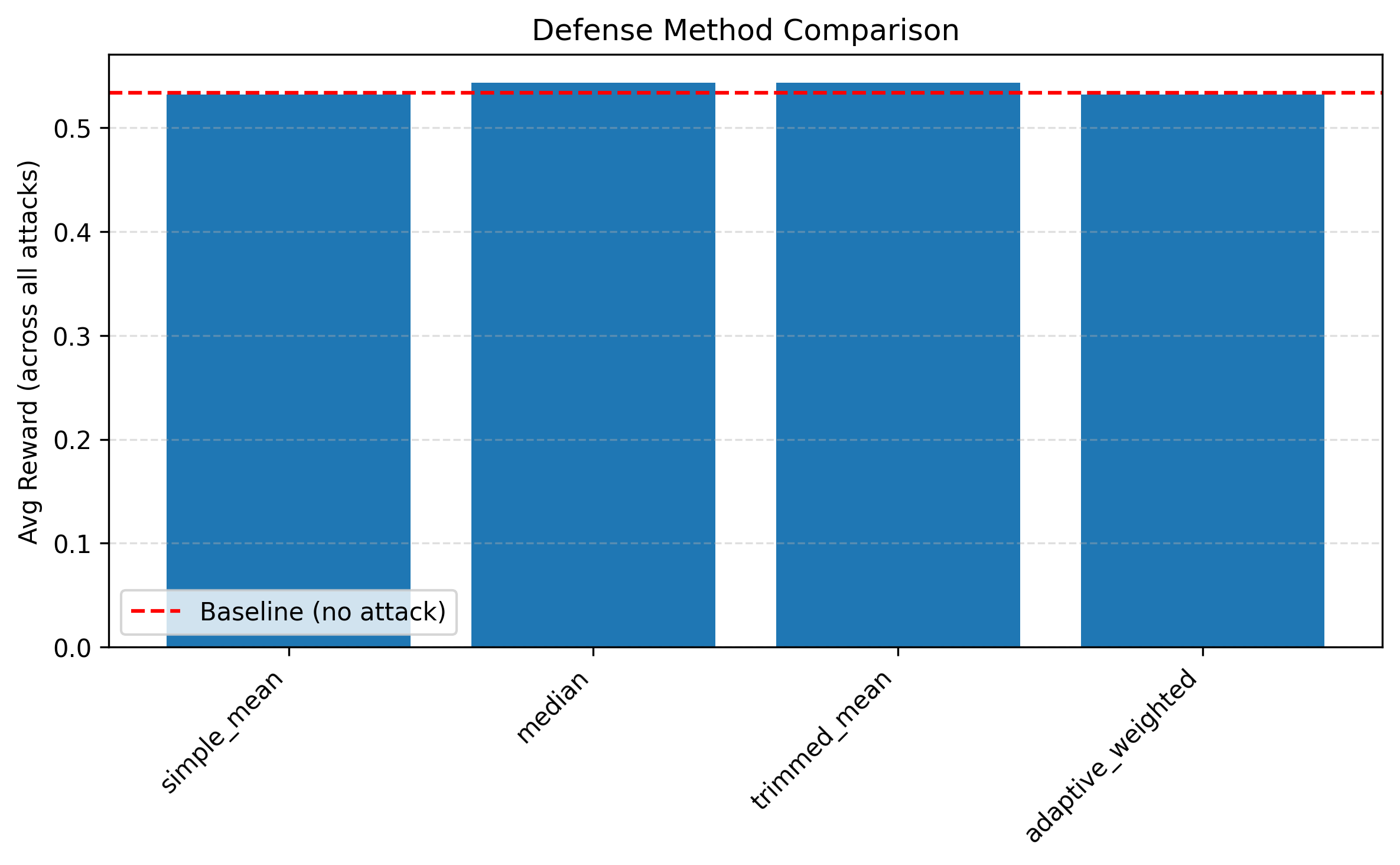}
  \caption{Defense summary in PoQ integration experiments. Robust aggregation and adaptive trust-weighting aim to reduce malicious evaluator influence.}
  \label{fig:defense-summary}
\end{figure}

\paragraph{Defense summary.}
Figure~\ref{fig:defense-summary} summarizes the defensive mechanisms and their qualitative effects.
The main message is that defenses which explicitly reduce outlier influence or adapt to evaluator reliability provide a strong foundation, but their effectiveness still depends on the underlying quality signal.

\begin{figure}[!htbp]
  \centering
  \includegraphics[width=0.7\linewidth]{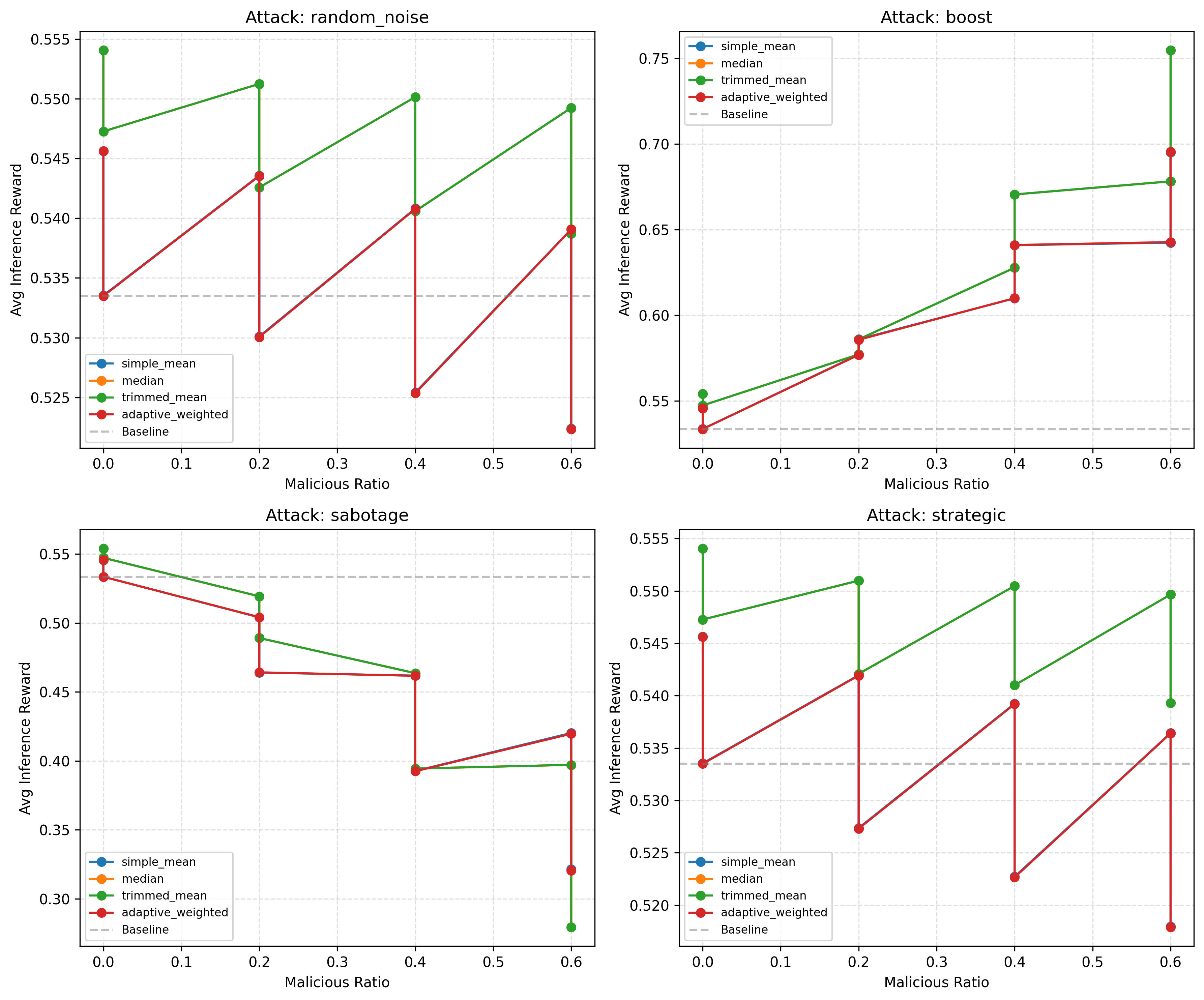}
  \caption{Adversarial defense comparison across attacker ratios and defense mechanisms. Calibrated composite scoring and robust PoQ defenses can be complementary.}
  \label{fig:adv-defense-comparison}
\end{figure}

\paragraph{Adversarial performance comparison.}
Figure~\ref{fig:adv-defense-comparison} provides a comparative view across attacker ratios and defense mechanisms.
We observe that robust aggregation and adaptive trust weighting mitigate attacks most effectively, and that calibrated composite scoring (i.e., removing unreliable dimensions identified in Section~\ref{sec:results}) can further stabilize reward outcomes by improving measurement alignment before aggregation.

\paragraph{Takeaway.}
PoQ provides a practical incentive backbone for decentralized inference, but it amplifies the importance of measurement.
Multi-dimensional scoring is most effective when paired with (i) reliability auditing and calibration at the dimension level (Section~\ref{sec:results}), and (ii) robust aggregation and adaptive trust mechanisms at the protocol level \citep{tian2026adaptiverobustpoq}.

\section{Discussion}
\label{sec:discussion}

Our results highlight a practical but often overlooked point: \emph{more signals do not automatically yield a better quality signal}.
This section distills implications for designing multi-dimensional scorers and deploying them inside PoQ-like incentive mechanisms.

\paragraph{Why multi-dimensional scoring can fail without calibration.}
Two dimensions that are intuitively appealing---query-output alignment and agreement/uncertainty---can reduce overall alignment when (i) the task objective differs from what the evaluator captures, or (ii) evaluator heterogeneity introduces directionality inversions.
This echoes broader findings that learned metrics must be validated on the target distribution and objective rather than assumed reliable \citep{sellam-etal-2020-bleurt,rei-etal-2020-comet,liang2023helm}.

\paragraph{Task dependence is the norm, not an edge case.}
We observe sharp differences across QA and summarization: alignment- and agreement-based signals can behave oppositely depending on task.
Summarization evaluation further requires factual consistency beyond semantic similarity \citep{kryscinski-etal-2020-evaluating,scialom-etal-2021-questeval,laban-etal-2022-summac}, which motivates task-aware calibration or modular activation (e.g., enable NLI-style consistency only when the task benefits from it).

\paragraph{Calibration strategies.}
Our study suggests three actionable calibration strategies:
(i) \emph{reliability auditing} (dimension-wise correlation and ablations),
(ii) \emph{task-wise weighting} (separate $\{w_k\}$ per task family or prompt cluster), and
(iii) \emph{gated composition} (enable a dimension only when it passes a reliability threshold on recent traffic).
These strategies preserve interpretability while avoiding overfitting to a single metric.

\paragraph{Interaction with PoQ defenses.}
Composite scoring and PoQ robustness mechanisms are complementary.
Even a calibrated composite can be undermined by malicious or unreliable evaluators; conversely, robust aggregation and trust weighting cannot fully fix a systematically misaligned signal \citep{tian2026adaptiverobustpoq}.
This mirrors lessons from Byzantine robustness: both \emph{estimator robustness} and \emph{signal validity} matter \citep{castro1999pbft,blanchard2017krum,pmlr-v80-yin18a,mhamdi2018hidden}.

\paragraph{Deployment guidance and monitoring.}
In production decentralized inference, we recommend:
(1) periodically re-estimating dimension reliability and updating weights,
(2) monitoring distribution shift in dimension outputs (e.g., drift in structure/length statistics),
and (3) keeping a ``safe fallback'' signal (e.g., a strong semantic evaluator) when reliability tests fail.
Cost-aware evaluation remains essential: cheaper dimensions can filter obvious failures, reserving expensive evaluators for borderline cases \citep{tian2025costawarepoq}.

\paragraph{Limitations.}
Our evaluation relies on reference signals that approximate human preference; no single reference fully defines quality across tasks.
Moreover, dimension instantiations (e.g., a specific NLI model) can materially affect results.
These limitations further reinforce the paper's message: multi-dimensional scoring should be treated as an auditable, continuously calibrated layer rather than a one-shot metric.

\section{Related Work}
\label{sec:related}

\paragraph{Decentralized inference and efficient serving.}
Collaborative inference systems demonstrate the feasibility of pooling resources across distributed participants \citep{borzunov-etal-2023-petals}.
At the same time, efficient LLM serving techniques show that memory and latency constraints remain fundamental, motivating cost-aware designs even before decentralization \citep{kwon2023pagedattention,dao2022flashattention}.

\paragraph{Verifiable computation and incentives.}
Cryptographic approaches to verifiable computation provide strong guarantees but are often costly for modern deep models, especially under low-latency serving constraints \citep{parno2013pinocchio,bensasson2014vonneumann}.
PoQ-style mechanisms instead use learned or statistical evaluation signals as a practical verification surrogate, enabling scalable incentives in decentralized inference \citep{tian2025costawarepoq,tian2026adaptiverobustpoq}.

\paragraph{Automatic evaluation and learned metrics.}
Classic overlap-based metrics such as BLEU and ROUGE remain widely used but are known to be brittle proxies for human judgment \citep{papineni-etal-2002-bleu,lin-2004-rouge}.
Embedding- and learned-metric approaches improve correlation in many settings \citep{zhang2020bertscore,zhao-etal-2019-moverscore,sellam-etal-2020-bleurt,rei-etal-2020-comet}, while task-specific evaluation (e.g., factual consistency for summarization) remains an active area \citep{kryscinski-etal-2020-evaluating,scialom-etal-2021-questeval,laban-etal-2022-summac}.
Our work complements this literature by emphasizing \emph{dimension reliability} and \emph{task dependence} in the context of decentralized incentives.

\paragraph{Holistic and preference-based evaluation.}
Holistic evaluation frameworks and LLM-as-a-judge protocols aim to capture broader capabilities beyond single metrics \citep{liang2023helm,zheng2023judging}.
Human-preference platforms such as Chatbot Arena provide scalable comparative judgments and associated ranking methodologies \citep{chiang2024chatbotarena,herbrich2007trueskill}.
We leverage these insights to motivate priors and to frame quality as multi-dimensional rather than monolithic.

\paragraph{Robust aggregation and Byzantine resilience.}
Robust consensus and learning under adversaries motivate median/trimmed strategies and adaptive trust mechanisms \citep{castro1999pbft,blanchard2017krum,pmlr-v80-yin18a,mhamdi2018hidden}.
Our prior work adapts these ideas to PoQ \citep{tian2026adaptiverobustpoq}, and this paper shows that robust aggregation is most effective when paired with a calibrated multi-dimensional quality signal.

\paragraph{Multi-dimensional quality models.}
Multi-dimensional quality assessment has a long history in language evaluation, including frameworks that separate quality into interpretable categories \citep{burchardt-2013-multidimensional}.
Our contribution is to operationalize a modular, auditable multi-dimensional scorer for decentralized inference and to study its reliability and integration with PoQ incentives.

\section{Conclusion}
\label{sec:conclusion}

We introduced a multi-dimensional quality scoring framework for decentralized LLM inference and studied how individual dimensions and their composites align with reference quality signals.
Our results show that multi-dimensional scoring is powerful but non-trivial: intuitive dimensions such as alignment and agreement can be task-dependent and even negatively aligned without calibration.
We therefore advocate treating quality scoring as an auditable and continuously calibrated layer.
Finally, we demonstrated that the composite score can be integrated as a drop-in quality signal for Proof of Quality, complementing cost-aware evaluation and robust aggregation mechanisms \citep{tian2025costawarepoq,tian2026adaptiverobustpoq}.

\clearpage
\bibliographystyle{plainnat}
\bibliography{references}

\end{document}